\documentclass[sigconf, authorversion=true, nonacm=true]{acmart}

\AtBeginDocument{%
  }

\usepackage{svg}
\usepackage{multirow}
\usepackage{balance} %KvL
\copyrightyear{2026}
\acmYear{2026}
\setcopyright{cc}
\setcctype{by}
\acmConference[ISWC '26]{Proceedings of the 2026 ACM International Symposium on Wearable Computers}{October 11--15, 2026}{Shanghai, China}
\acmBooktitle{Proceedings of the 2026 ACM International Symposium on Wearable Computers (ISWC '26), October 11--15, 2026, Shanghai, China}
\acmDOI{10.1145/3830727.3834810}
\acmISBN{979-8-4007-2872-3/2026/11}

\begin{document}

\title{DeepConvContext: A Multi-Scale Approach to Timeseries Classification in Human Activity Recognition}

\author{Marius Bock}
\email{bock@iai.uni-bonn.de}
\orcid{0000-0001-7401-928X}
\affiliation{
  \institution{University of Bonn}
  \institution{Lamarr Institute for ML and AI}
  \city{Bonn}
  \country{Germany}
}
\author{Juergen Gall}
\email{gall@iai.uni-bonn.de}
\orcid{0000-0002-9447-3399}
\affiliation{
  \institution{University of Bonn}
  \institution{Lamarr Institute for ML and AI}
  \city{Bonn}
  \country{Germany}
}

\author{Michael Moeller}
\email{michael.moeller@uni-siegen.de}
\orcid{0000-0002-0492-6527}
\affiliation{
  \institution{University of Siegen}
  \city{Siegen}
  \country{Germany}
}

\author{Kristof Van Laerhoven}
\email{kvl@eti.uni-siegen.de}
\orcid{0000-0001-5296-5347}
\affiliation{
  \institution{University of Siegen}
  \city{Siegen}
  \country{Germany}
}

\renewcommand{\shortauthors}{Marius Bock, Juergen Gall, Michael Moeller, Kristof Van Laerhoven}

\begin{abstract}
    Despite recognized limitations in modeling long-range temporal dependencies, Human Activity Recognition (HAR) has traditionally relied on a sliding window approach to segment labeled datasets. Deep learning models like the DeepConvLSTM typically classify each window independently, restricting learnable temporal context to within-window information and producing fragmented, temporally incoherent activity timelines. To address this constraint, we propose \textit{DeepConvContext}, a multi-scale time series classification framework for HAR. Drawing inspiration from the vision-based Temporal Action Localization community, DeepConvContext models both intra- and inter-window temporal patterns separately by processing sequences of time-ordered windows. Across six widely-used HAR benchmarks, DeepConvContext achieves an average 5\% improvement in F1-score and up to 18-point improvement in mAP over related approaches, while achieving latency and throughput comparable to prior methods that extend temporal context through hidden state propagation across batches. Our quantitative and qualitative analysis underline the importance of inter-window learning and show how it produces more coherent activity segments even in online prediction scenarios. Code to reproduce our experiments is publicly available via \url{www.github.com/mariusbock/deepconvcontext}.
\end{abstract}

\begin{CCSXML}
<ccs2012>
<concept>
<concept_id>10003120.10003138.10003142</concept_id>
<concept_desc>Human-centered computing~Ubiquitous and mobile computing design and evaluation methods</concept_desc>
<concept_significance>500</concept_significance>
</concept>
<concept>
<concept_id>10010147.10010257.10010293.10010294</concept_id>
<concept_desc>Computing methodologies~Neural networks</concept_desc>
<concept_significance>500</concept_significance>
</concept>
</ccs2012>
\end{CCSXML}

\ccsdesc[500]{Human-centered computing~Ubiquitous and mobile computing design and evaluation methods}
\ccsdesc[500]{Computing methodologies~Neural networks}

\keywords{Human Activity Recognition, Deep Learning, CNN-LSTMs}

\begin{teaserfigure}
\centering
  \includegraphics[width=1.0\textwidth]{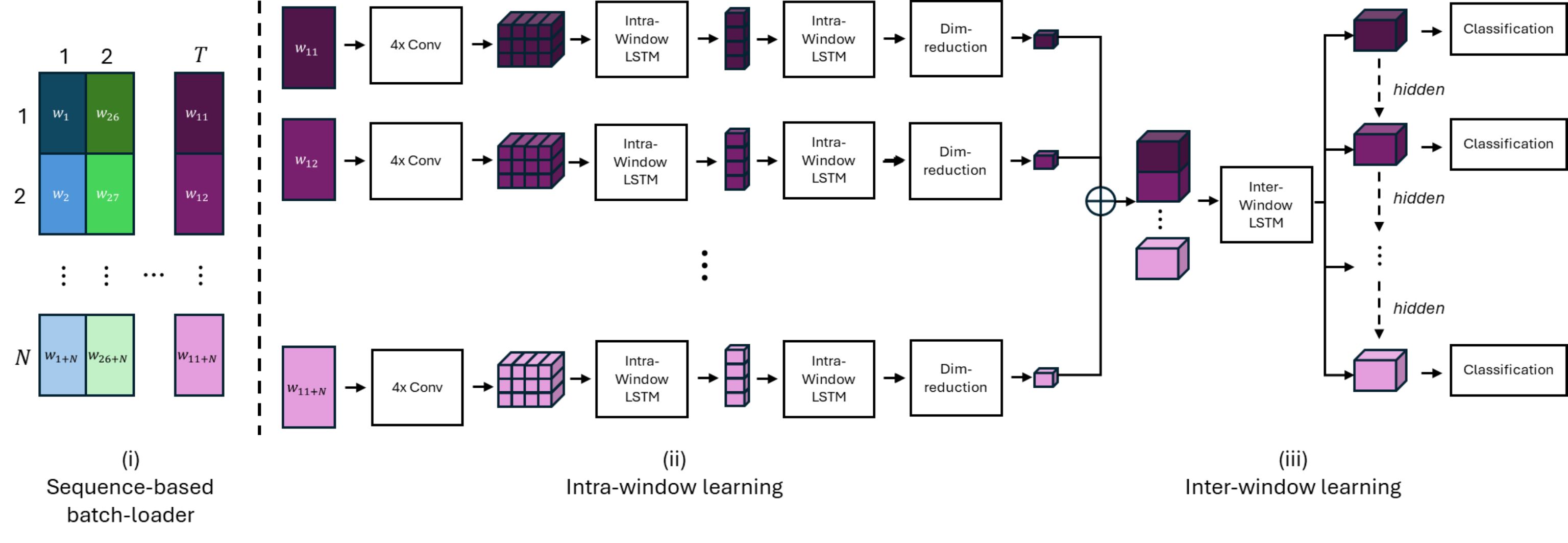}\vspace{-14pt}
  \caption{\textit{Overview of DeepConvContext.} Our architecture uses a multi-scale approach for sliding window activity classification: Using a sequence-based shuffled batch loader during training (i), windows are individually processed by a \textit{DeepConvLSTM}-like feature extraction (intra-window learning, ii). The resulting sequence of feature vectors of all windows (iii) are then passed to a second LSTM, to learn inter-window temporal features and are then individually classified. By preserving hidden states across windows, DeepConvContext generalises to arbitrary-length test sequences, enabling online prediction.}
    \Description{}
  \label{fig:teaser}
\end{teaserfigure}

\maketitle
\vspace{-10pt}
\section{Introduction}
\label{sec:intro}

Inertial-based Human Activity Recognition (HAR) in wearable computing research has, over the past two decades, traditionally relied on a sliding window-based classification approach \cite{bullingTutorialHumanActivity2014}. Enabling near real-time prediction, this method divides continuous sensor data into overlapping windows, each of which is independently classified by a machine learning model, such as the DeepConvLSTM \cite{ordonezDeepConvolutionalLSTM2016}. While subsequent research has sought to enhance models' capabilities through the integration of mechanisms like attention \cite{zhouTinyHARLightweightDeep2022} to better model temporal patterns, these models remain fundamentally constrained by their dependence on sliding windows, restricting broader temporal understanding across windows \cite{bullingTutorialHumanActivity2014}.

Knowing these temporal limitations of window-based classification, only limited research has aimed to address these with alterations to training procedures of HAR models \cite{pellattCausalBatchSolvingComplexity2020, hiremathRoleContextLength2021, shaoIsolatedFramesEnhancing2024}. Most prominently, \citet{pellattCausalBatchSolvingComplexity2020} suggested to extend reasoning of LSTMs used within model architectures by making positions among batches temporally connected during training and passing along hidden states between batches. 
While construction of these so called \textit{CausalBatches} showed to significantly improve classification results, giving models a larger temporal horizon while still using small-sized sliding windows, the LSTM used within these models are tasked to model both local and global temporal structure, yet are optimized on batch-wise gradients derived only from within-batch windows and hence gradients that capture only local patterns. We argue in this work that this mismatch between what the model is asked to learn and what the training signal actually encodes is a fundamental limitation of the CausalBatch approach.

Recently, vision-based Temporal Action Localization (TAL) models have been successfully applied to inertial data and shown to outperform established HAR architectures such as DeepConvLSTM across various datasets \cite{bockTemporalActionLocalization2024}. While these TAL models are designed for offline timeline reconstruction, we argue that their core design choice of separating local feature extraction from global temporal reasoning into distinct model components naturally maps onto the intra- and inter-window structure of sliding-window HAR as well as existing models. With human behaviour evolving naturally through local and global patterns, the TAL community has identified that temporal understanding of the past (as well as the future) leading up to a current timestamp that is sought to be classified, is crucial for reliably detecting different types of activities \cite{shiTriDetTemporalAction2023}. 

We thus introduce \textit{DeepConvContext}, a novel inertial-based architecture inspired by TAL methodologies, which is is specifically designed to model both short-term (intra-window) and long-term (inter-window) temporal dependencies across sequences of sliding windows of inertial data. 
Our contributions are three-fold:
\begin{enumerate}
    \item We propose \textit{DeepConvContext}, a novel architecture which effectively uses sequence-based batches for multi-scale recognition via an inter-window LSTM and achieves comparable latency and throughput to related benchmarks, despite its additional module. 
    \item An extensive evaluation on six widely-used HAR benchmark datasets shows that DeepConvContext outperforms the standard DeepConvLSTM by up to 21\% and CausalBatch by up to 10\% in F1-score, while producing the highest mAP scores across all experiments, improving over related baselines by up to 18 points in an online prediction scenario.
    \item During ablation experiments we demonstrate that LSTMs consistently outperform attention- and  transformer-based inter-window mechanisms by up to 5\% in F1-score across all benchmarks, suggesting that the sequential, locally-ordered inductive bias of LSTMs is better suited to HAR than the global, order-agnostic structure of self-attention.
\end{enumerate}

\vspace{-10pt}

\section{Related Work}
\label{sec:related}

With the rise of deep learning, researchers have adapted convolutional neural networks (CNNs) for use on time series sensor data, eliminating the need for manual feature extraction that previously relied on domain expert knowledge. Known as the \textit{DeepConvLSTM}, \citet{ordonezDeepConvolutionalLSTM2016} extended this approach by incorporating recurrent layers into inertial-based architectures, enabling them to capture temporal relationships between convolutional features. Despite significant progress in improving methods for extracting temporal information, e.g. via attention mechanisms \cite{abedinAttendDiscriminateStateOfTheArt2021, zhouTinyHARLightweightDeep2022, murahariAttentionModelsHuman2018}, these architectures rely on a sliding window approach, where each window of sensor data is processed independently to predict its label. As detailed by \citet{bullingTutorialHumanActivity2014}, this fixed window size limits the model's ability to capture inter-window temporal dependencies, making the choice of window length a critical factor in HAR performance.

Research has therefore emphasized the need for models that can capture both local and global temporal patterns to overcome the limitations of the sliding window approach \cite{hiremathRoleContextLength2021, hammerlaLetsNotStick2015, shaoIsolatedFramesEnhancing2024}.  \citet{hiremathRoleContextLength2021} explored the impact of larger context lengths on HAR performance, aggregating convolutional features using larger kernels before passing them through an LSTM. \textit{CausalBatch}~\cite{pellattCausalBatchSolvingComplexity2020} extends the LSTM's temporal horizon by making windows within a batch causally dependent on those in neighbouring batches, allowing LSTM states to persist across batch boundaries without increasing the window size. The Shallow DeepConvLSTM~\cite{bockImprovingDeepLearning2021}, even though it was primarily proposed as a more lightweight alternative to the original DeepConvLSTM, inherently introduces an inter-window learning effect by applying the LSTM across the batch sequence rather than within individual windows. DeepConvContext builds on the insights of both approaches by introducing a dedicated second LSTM explicitly designed for inter-window context modeling, keeping local and global temporal learning structurally separate, making inter-window reasoning an explicit, optimized objective.

\section{Methodology}
\label{sec:methodolgy}

\subsection{DeepConvContext}
\label{subsec:architecture}
Figure~\ref{fig:teaser} illustrates the proposed \textit{DeepConvContext} architecture, which combines three key design choices.
\begin{enumerate}
\item \textbf{Sequence-based batch loading.} Following \citet{shaoIsolatedFramesEnhancing2024}, we employ a batch loader that, at each iteration, randomly selects a participant and a starting point within their input time series, and extracts a contiguous sequence of $B$ windows from that point. Unlike \textit{CausalBatch}~\cite{pellattCausalBatchSolvingComplexity2020}, which connects hidden states across otherwise unordered batches, this preserves temporal consistency among batch elements directly, enabling the model to learn inter-window patterns from more informative gradient updates. To avoid overfitting, the sampling start point is randomized per batch, and all LSTM hidden states are reset between batches.

\item \textbf{Intra-window feature extraction.} Each window in the input sequence is processed independently by a feature extractor that is identical to the original DeepConvLSTM \cite{ordonezDeepConvolutionalLSTM2016}. LSTM hidden states are reset between windows, ensuring that this module captures only local, within-window temporal structure. The output of each window's extractor is summarized into a fixed-size feature vector of dimension $1 \times LSTM_h$, where $LSTM_h$ denotes the number of hidden units in the final intra-window LSTM layer.

\item \textbf{Inter-window context modeling.} Lastly, the sequence of $B$ window-level feature vectors is passed to a second, context-based LSTM that models temporal dependencies across windows. Its output sequence is passed through a dropout layer and a classification layer to produce per-window predictions. Because each window is processed by both the intra- and inter-window LSTMs, the final output vectors encode both local temporal structure and each window's temporal relationship to earlier windows in the sequence.    
 
\end{enumerate}

\subsection{Datasets}
\label{subsec:datasets}

We base our experiments on six widely used HAR datasets: the Wetlab \cite{schollWearablesWetLab2015}, WEAR \cite{bockWEAROutdoorSports2024}, SBHAR \cite{reyes-ortizTransitionAwareHumanActivity2016}, RWHAR \cite{sztylerOnBodyLocalizationWearable2016}, Opportunity \cite{roggenCollectingComplexActivity2010}, and Hang-Time datasets \cite{hoelzemannHangTimeHARBenchmark2023}. This selection provides a diverse set of prediction scenarios, each posing distinct challenges for activity recognition models to address. Among these challenges are variations in the number of participants and activity classes, the presence of a NULL-class \cite{bockWEAROutdoorSports2024, schollWearablesWetLab2015, hoelzemannHangTimeHARBenchmark2023, roggenCollectingComplexActivity2010}, and the inclusion of complex, short-duration, transitional, and locomotion activities \cite{bockWEAROutdoorSports2024, hoelzemannHangTimeHARBenchmark2023, roggenCollectingComplexActivity2010, schollWearablesWetLab2015, reyes-ortizTransitionAwareHumanActivity2016, sztylerOnBodyLocalizationWearable2016}. Additionally, the datasets vary in sensor configurations, with several collected in multi-sensor environments \cite{bockWEAROutdoorSports2024, roggenCollectingComplexActivity2010, sztylerOnBodyLocalizationWearable2016}.

\subsection{Baselines \& Architecture Variants}
\label{subsec:variants}
We compare DeepConvContext against three baselines: (1) a classic DeepConvLSTM trained on randomly sampled windows, (2) a DeepConvLSTM trained using \textit{CausalBatch}~\cite{pellattCausalBatchSolvingComplexity2020}, and (3) the Shallow DeepConvLSTM~\cite{bockImprovingDeepLearning2021}. Specifically, each of these baselines models different temporal structures: shuffling restricts (1) to intra-window context only, while (2) and (3) extend temporal context during training by propagating the LSTM hidden state between batches, resetting every $l=100$ batches in (2) and across batch elements in (3). During testing, all baselines and the proposed DeepConvContext are applied online, predicting participant timelines batch by batch with hidden states carried across the full test timeline, ensuring minimal latency.

LSTMs, though effective at modeling local temporal dependencies, are known to struggle with long-range relationships as the temporal distance between relevant elements grows \cite{bengioLearningLongtermDependencies1994, pascanuDifficultyTrainingRecurrent2013}. Self-attention mechanisms can capture dependencies between any two sequence elements regardless of temporal distance, and have driven notable performance gains in inertial sensor-based architectures such as Attend-and-Discriminate \cite{abedinAttendDiscriminateStateOfTheArt2021}, making them a natural candidate for inter-window context modeling. However, unlike recurrent architectures, attention-based models lack a natural sequential state that can be updated incrementally, making per-window online inference computationally expensive. This limits their suitability for streaming deployment compared to LSTMs, which process inputs one step at a time with constant-cost state updates. 
We therefore include both multi-head attention and transformer layers as controlled drop-in replacements for the inter-window LSTM in our ablation study, rather than as primary design choices. To ensure comparability to unidirectional LSTMs, we restrict both attention-based variants to their causal formulations. Nevertheless, since activity recognition can benefit from future context at prediction time, we also report results for bidirectional variants of all three inter-window mechanisms (LSTM, attention, and transformer). Note that bidirectional LSTMs, same as attention-based variants, require access to a full batch input sequence before inference, making them less suited for online activity recognition purposes.

\subsection{Training}
\label{subsec:training}
For each experiment, we perform Leave-One-Subject-Out (LOSO) cross-validation, where each participant in the dataset is used as the validation set once, while all other participants are used for training. Each experiment is repeated three times with varying random seeds. Averaging these three runs, we report two key metrics for each experiment: the class-averaged macro F1-score and the mean Average Precision (mAP), both averaged across all validation splits (i.e., participants). The mAP, which evaluates the overlap between predicted activity segments and ground truth annotations, is computed across five temporal Intersection-over-Union (tIoU) thresholds (0.3, 0.4, 0.5, 0.6, and 0.7), following the evaluation conventions established in the TAL literature \cite{shiTriDetTemporalAction2023}. We report the average across these five thresholds as the final mAP metric for each experiment. 
To ensure that both metrics (F1 and mAP) are computed on the same temporal resolution, we convert the windowed predictions of each model back to per-sample predictions by unwindowing them. 
The final reported values per dataset are the averages of the tIoU-averaged mAP and class-averaged F1-score across all validation subjects. All model architectures are trained using a weighted cross-entropy loss and the Adam optimizer, with a learning rate of $1e^{-4}$, weight decay of $1e^{-6}$. Each model is trained for 30 epochs, using a fixed learning rate schedule that multiplies the learning rate by a factor of 0.9 every 10 epochs. 
All architectures evaluated in this benchmark use four convolutional layers with 64 filters, each of size $9 \times 1$, except for the Opportunity dataset, where a smaller filter size of $5 \times 1$ is used due to the lower sampling rate of the input signal. As suggested by \citet{bockImprovingDeepLearning2021}, LSTMs use a single layer with a hidden size of 128 (see appendix for comparison to two layered variants). In ablation experiments where LSTMs are replaced with transformers or multi-head self-attention mechanisms, we use four attention heads with a hidden dimension of 768 and no dropout. The transformer-based variants use two layers, 4 heads and a smaller hidden dimension of 256. Same as Ordóñez and Roggen \cite{ordonezDeepConvolutionalLSTM2016}, embeddings get passed through a dropout layer with a probability of 0.5 before being passed to the classifier. 

\section{Results}
\label{sec:results}

\begin{figure*}
    \centering
    \includegraphics[width=0.8\linewidth]{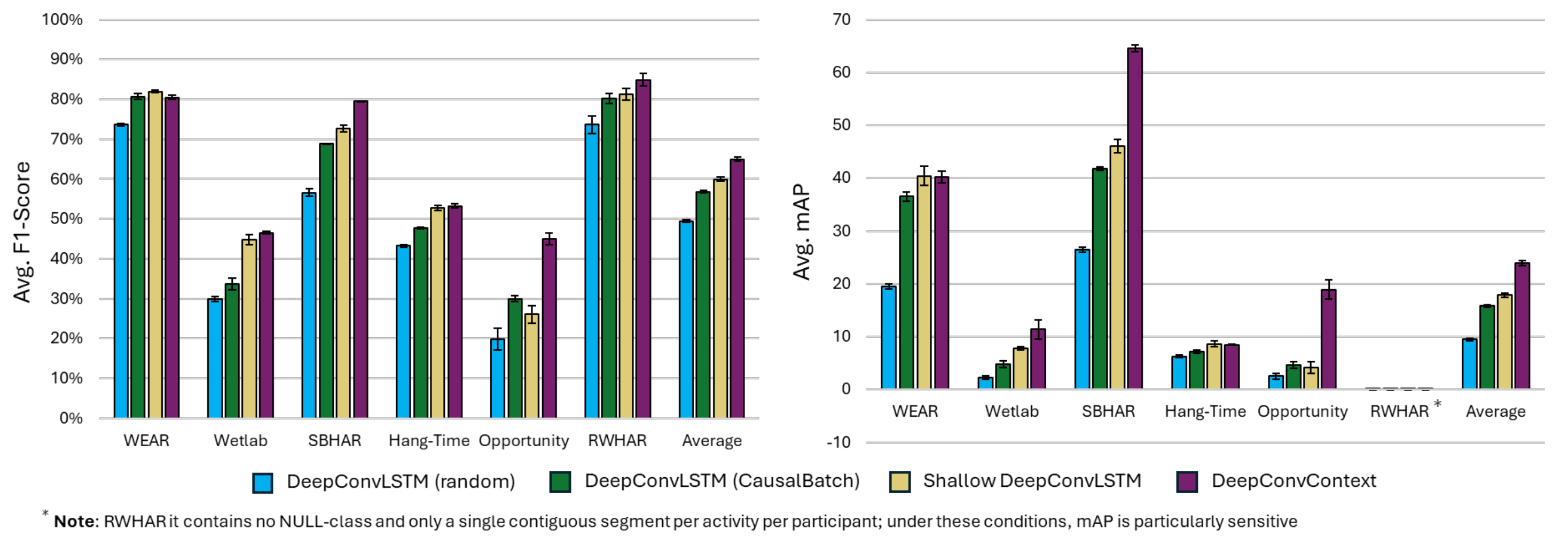}\vspace{-12pt}
    \caption{Average F1-score and mAP results of the DeepConvLSTM \cite{ordonezDeepConvolutionalLSTM2016}, Shallow DeepConvLSTM \cite{bockImprovingDeepLearning2021} and proposed DeepConvContext applied to the investigated datasets. One can see that the DeepConvContext combines strengths of both architectures and improves upon results across all datasets, providing the highest F1-score and mAP.}
    \Description{}
    \label{fig:overview}
\end{figure*}

\subsection{Benchmark analysis}

Figure~\ref{fig:overview} presents results for the standard DeepConvLSTM \cite{ordonezDeepConvolutionalLSTM2016}, a CausalBatch-trained DeepConvLSTM \cite{pellattCausalBatchSolvingComplexity2020}, the Shallow DeepConvLSTM \cite{bockImprovingDeepLearning2021}, and the proposed DeepConvContext, all using a single LSTM layer per module. On average, the multi-scale DeepConvContext achieves the highest prediction performance, with an average F1-score of 64.96\%, compared to 49.46\% for the standard, 56.84\% for the CausalBatch-trained and 59.92\% for the Shallow DeepConvLSTM, across the six evaluated datasets. Furthermore, as DeepConvContext leverages both intra- and inter-window context, it produces the highest mAP scores across all experiments, outperforming the Shallow DeepConvLSTM by as much as 18 score points for the SBHAR and 14 score points for the Opportunity dataset.

\begin{figure*}
    \centering
    \includegraphics[width=1.0\linewidth]{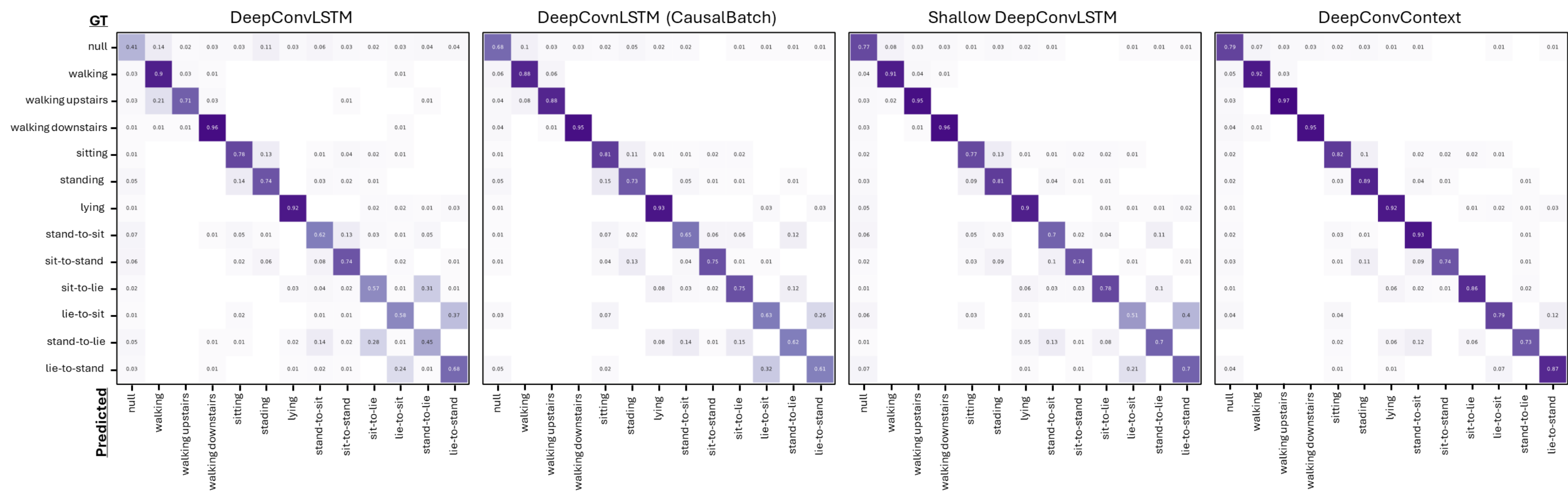}\vspace{-12pt}
    \caption{Per-class confusion matrices of the DeepConvLSTM, Shallow DeepConvLSTM and DeepConvContext being applied to the SBHAR dataset using LOSO cross-validation. DeepConvContext improves upon these bechmarks, with specifically transition classes such as \textit{sit-to-stand} being more reliably detected.}
    \label{fig:confmats}
    \Description{}
\end{figure*}

\begin{figure}
    \centering
    \includegraphics[width=1.0\linewidth]{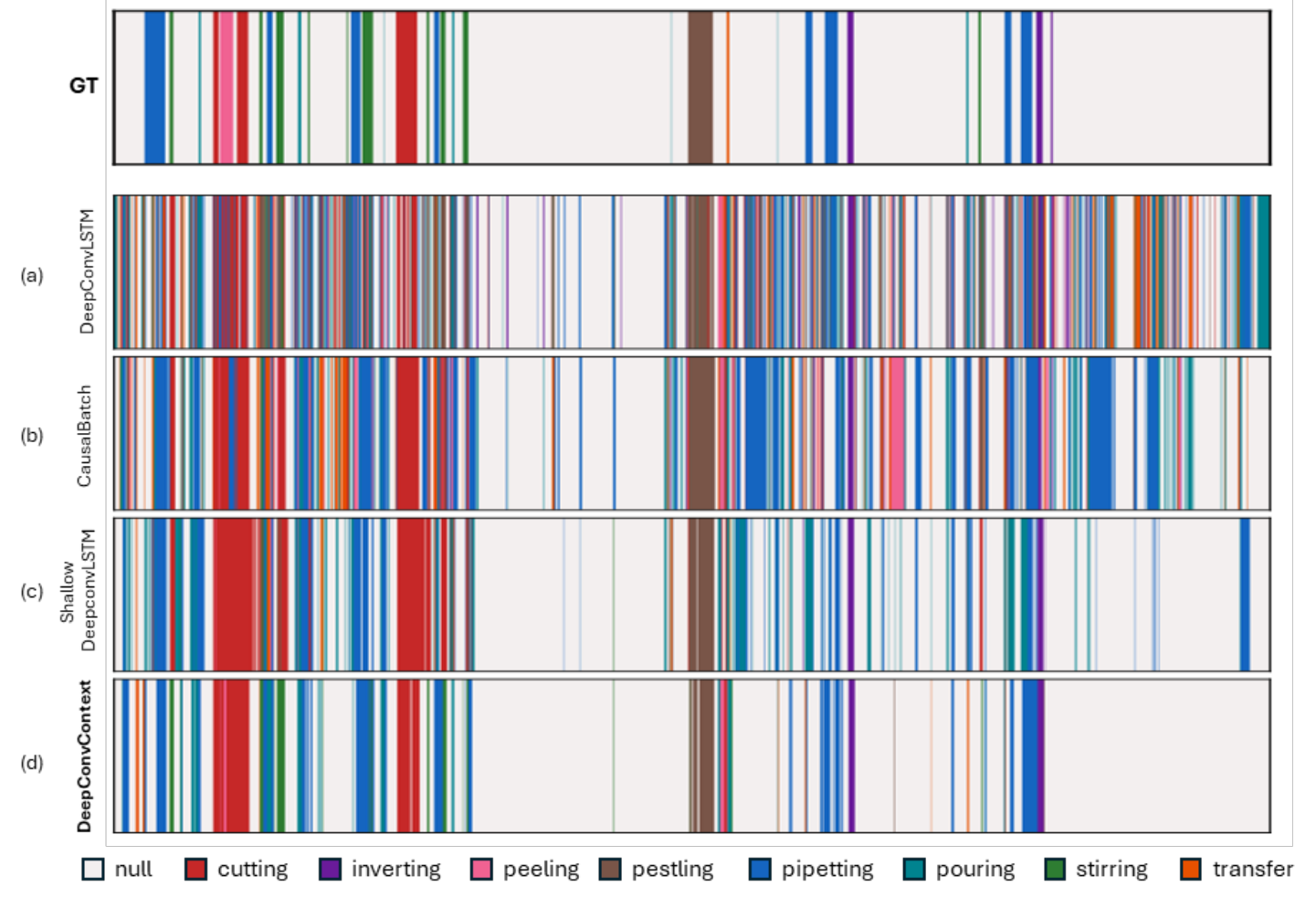}
    \caption{Color-coded visualization of a sample participant from the Wetlab dataset reconstructed by (a) classic DeepConvLSTM, (b) CausalBatch, (c) Shallow DeepConvLSTM and (d) our proposed DeepConvContext. One can see that DeepConvContext most effectively recognizes and reconstructs ground truth (GT) segments.}%\vspace{-10pt}}
    \label{fig:color_viz_wetlab}
    \Description{}
        \vspace{-10pt}

\end{figure}

Although all models propagate LSTM hidden states across predictions at test time, the standard DeepConvLSTM, trained without inter-window supervision, consistently produces the least coherent segments, as reflected in its substantially lower mAP across all datasets. While \citet{pellattCausalBatchSolvingComplexity2020} claim to address the limited temporal reasoning of the DeepConvLSTM, both mAP results and qualitative segment visualizations in Figure~\ref{fig:color_viz_wetlab} show that sequence-based learning, as used in the Shallow DeepConvLSTM and DeepConvContext, outperforms CausalBatch in both prediction accuracy and timeline reconstruction quality. As also seen in Figure~\ref{fig:color_viz_wetlab}, DeepConvContext shows less frequent misclassifications of NULL windows into activity classes, which we hypothesize is due to its dedicated inter-window LSTM that helps the model leverage surrounding temporal context to disambiguate locally activity-like but isolated windows. We attribute this to the stronger inter-window training signal produced by gradients computed over temporally ordered sequences, as opposed to the within-batch gradients of CausalBatch which capture only local structure. All in all, the DeepConvContext effectively combines the strengths of both intra- and inter-window learning, maintaining high predictive performance on all activity type, as shown in the confusion matrices in Figure~\ref{fig:confmats}.

\subsection{Ablation experiments}

\paragraph{Impact of Context Length}

With the DeepConvContext architecture using temporal sequences as batches, applied batch sizes during training become an important tuneable hyperparameter as they influence the learnable inter-window context. To measure sensitivity of our proposed approach to the applied training batch size, Table~\ref{tab:batchsize} shows results of applying smaller (25 and 50) as well as a larger training batch size (200). Results show that DeepConvContext remains largely stable across all tested context lengths. Notably, even with as few as 25 windows (corresponding to 12.5 seconds of data) the model retains competitive performance across most datasets, suggesting that inter-window learning is effective even under the short-buffer constraints typical of wearable deployment. Nevertheless, datasets with complex or long-duration activities, such as RWHAR and Opportunity, show consistent gains with longer context, indicating that the optimal batch size is activity-type dependent and represents a tunable deployment parameter.

\begin{table}
    \centering
    \small
    \caption{Average F1-score and mAP results of DeepConvContext using varying train sequence batch sizes (25, 50 and 200). We report results for the WEAR \cite{bockWEAROutdoorSports2024}, Wetlab \cite{schollWearablesWetLab2015}, Hang-Time \cite{hoelzemannHangTimeHARBenchmark2023}, RWHAR \cite{sztylerOnBodyLocalizationWearable2016}, Opportunity \cite{roggenCollectingComplexActivity2010} and SBHAR \cite{reyes-ortizTransitionAwareHumanActivity2016}. Results show that DeepConvContext remains largely stable across context lengths, with longer sequences yielding modest gains on datasets with complex or long-duration activities.}
    \label{tab:batchsize}
    \begin{tabular}{l|cc|cc|cc}
    $D$                                                  & \multicolumn{2}{c}{25} & \multicolumn{2}{c}{50} & \multicolumn{2}{c}{200} \\
                                                         & F1  & mAP & F1  & mAP & F1  & mAP \\ \hline
    \cite{bockWEAROutdoorSports2024}                     & 79.14\% & 36.43 & 80.78\% & 39.38 & 80.22\% & 41.07 \\
    \cite{schollWearablesWetLab2015}                     & 42.04\% & 9.50  & 45.39\% & 10.87 & 46.63\% & 10.83 \\
    \cite{hoelzemannHangTimeHARBenchmark2023}            & 53.97\% & 9.13  & 52.84\% & 8.60 & 52.67\% & 8.54 \\
    \cite{sztylerOnBodyLocalizationWearable2016}         & 80.89\% & 0.0   & 85.28\% & 0.01 & 88.62\% & 0.01 \\
    \cite{roggenCollectingComplexActivity2010}           & 39.07\% & 12.01 & 43.84\% & 17.47 & 46.71\% & 20.88 \\
    \cite{reyes-ortizTransitionAwareHumanActivity2016}   & 74.82\% & 53.47 & 77.36\% & 56.63 & 76.17\% & 62.96 \\ \hline
    \textit{Avg}                                         & 61.65\% & 20.09 & 64.25\% & 22.16 & 65.17\% & 24.05%
    \end{tabular}
        \vspace{-10pt}

\end{table}

\paragraph{LSTMs vs. Attention}

As detailed in Section~\ref{subsec:variants}, we replace the newly introduced second context-based LSTM in the DeepConvContext architecture with either a multi-head attention block or a transformer module. Results in Table~\ref{tab:attentionvslstms} show that LSTMs outperform attention-based mechanisms consistently across the selected benchmark datasets by as much as 5\% in F1-score. With LSTMs being sequential by design, LSTM-based results further produce higher mAP scores, as they allow passing of hidden states among batches and are thus not dependent on applying the same batch size during training and testing. Lastly, while attention mechanisms allow features to interact across the entire input sequence, they typically exhibit a reduced sensitivity to the natural order of events in time and rely on positional encodings to provide temporal structure. Although we included such encodings in our models, the relatively lower performance of the attention-based variants suggests that LSTMs may be more effective at capturing temporal relationships among windows. 

\begin{table}
    \centering
    \small
    \caption{Average F1-score results of DeepConvContext using (unidirectional or bidirectional) LSTM, attention or a Transformer to learn inter-window context. We report results for WEAR \cite{bockWEAROutdoorSports2024}, Wetlab \cite{schollWearablesWetLab2015}, Hang-Time \cite{hoelzemannHangTimeHARBenchmark2023}, RWHAR \cite{sztylerOnBodyLocalizationWearable2016}, Opportunity \cite{roggenCollectingComplexActivity2010} and SBHAR \cite{reyes-ortizTransitionAwareHumanActivity2016}. LSTMs significantly outperform both attention and transformers. Best unidirectional and bidirectional results per dataset are \underline{underlined}.}
    \label{tab:attentionvslstms}
    \begin{tabular}{l|ccc|ccc}
    $D$                                                  & \multicolumn{3}{c}{Unidirectional}      & \multicolumn{3}{c}{Bidirectional} \\
                                                         & LSTM                & Att.    & Tran.   & LSTM                & Att.    & Tran. \\ \hline
    \cite{schollWearablesWetLab2015}                     & \underline{46.52\%} & 36.45\% & 35.94\% & \underline{53.63\%} & 44.21\% & 39.79\% \\
    \cite{bockWEAROutdoorSports2024}                     & \underline{80.50\%} & 79.21\% & 78.53\% & \underline{83.24\%} & 82.24\% & 80.83\% \\
    \cite{hoelzemannHangTimeHARBenchmark2023}            & \underline{53.27\%} & 48.81\% & 48.82\% & \underline{59.12\%} & 53.84\% & 50.55\% \\
    \cite{sztylerOnBodyLocalizationWearable2016}         & \underline{84.92\%} & 83.54\% & 83.10\% & 88.91\% & 88.31\% & \underline{88.93\%} \\ 
    \cite{roggenCollectingComplexActivity2010}           & \underline{45.06\%} & 39.18\% & 41.42\% & \underline{50.58\%} & 35.52\% & 37.90\% \\
    \cite{reyes-ortizTransitionAwareHumanActivity2016}   & \underline{79.48\%} & 69.41\% & 71.34\% & \underline{80.22\%} & 73.71\% & 75.87\% \\ \hline
    \textit{Avg}                                         & \underline{64.96\%} & 59.43\% & 59.86\% & \underline{69.28\%} & 62.97\% & 62.31\% 
    \end{tabular}
     \vspace{-10pt}
\end{table}

\paragraph{Complexity and Inference Efficiency}

Table~\ref{tab:complexity} compares the model complexity and inference efficiency of the proposed DeepConvContext with its variants as well as predecessors. Compared to CausalBatch, which maintains $B$ separate hidden states across batch boundaries during training, DeepConvContext processes a single temporally ordered sequence per batch. %This reduces GPU memory consumption from 165.5\,MB to 123.0\,MB, a 27\% reduction. 
By passing only a compressed, fixed-size feature vector per window to the inter-window LSTM DeepConvContext achieves a memory footprint comparable to the Shallow DeepConvLSTM (123.0\,MB vs. 112.5\,MB) and is comparable in latency (0.96\,ms vs. 0.53\,ms for CausalBatch) and throughput (105k vs. 188k samples per second) to CausalBatch while having significantly more learnable parameters.

\begin{table}[t]
    \centering
    \small
    \caption{Model complexity and inference efficiency of CausalBatch, Shallow DeepConvLSTM (Shallow~D.), and DeepConvContext (DCC) variants, all with 1-layer LSTMs, on a single NVIDIA GeForce RTX 3090. Attention-based variants refer to their non-causal variants. We report parameters ($\rho$), GPU memory, latency (mean\,$\pm$\,std, batch size\,=\,100), and throughput (sps). Though having almost three times as many parameters, DCC remains comparable in memory and latency to CausalBatch.}
    \setlength{\tabcolsep}{5pt}
    \begin{tabular}{l r r r r}
        Architecture & $\rho$ & Memory & Latency (ms)$\downarrow$ & (sps)$\uparrow$ \\
        \midrule
        CausalBatch   & 278k & 114.4\,MB & $0.53 \pm 0.03$ & 188k \\
        Shallow D.    & 278k & 112.5\,MB & $4.18 \pm 0.11$ & 24k \\
        \midrule
        DCC (LSTM)    & 704k & 123.0\,MB & $0.96 \pm 0.10$ & 105k \\
        DCC (Bi-LSTM) & 969k & 126.0\,MB & $1.39 \pm 0.16$ &  72k \\
        DCC (Att.)    & 639k & 121.1\,MB & $1.08 \pm 0.02$ &  93k \\
        DCC (Tran.)   & 837k & 124.1\,MB & $1.15 \pm 0.05$ &  87k \\
        \bottomrule
    \end{tabular}
    \label{tab:complexity}
    \vspace{-10pt}
\end{table}

\paragraph{Revisiting shallow LSTMs}

Using a similar approach as \cite{bockImprovingDeepLearning2021}, we compare investigated baselines, and DeepConvContext using 1-layered and 2-layered LSTMs. Results (see Table~\ref{tab:shallowvsdeep}) show that on average single-layer LSTMs outperform their two-layer counterparts across all tested datasets and models, with the exception being our tested baselines applied to the Opportunity dataset as well as the DeepConvContext applied to the RWHAR dataset.

\begin{table}
    \centering
    \small
    \caption{Average F1-score results of DeepConvLSTM (DCL), a CausalBatch-trained DeepConvLSTM, Shallow DeepConvLSTM (Shallow D.) and proposed DeepConvContext (DCC) using 1-layered (1-L) or 2-layered LSTMs (2-L) within the architectures. We report results for  WEAR \cite{bockWEAROutdoorSports2024}, Wetlab \cite{schollWearablesWetLab2015}, Hang-Time \cite{hoelzemannHangTimeHARBenchmark2023}, RWHAR \cite{sztylerOnBodyLocalizationWearable2016}, Opportunity \cite{roggenCollectingComplexActivity2010} and SBHAR \cite{reyes-ortizTransitionAwareHumanActivity2016}. On average one-layered LSTMs outperform 2-layered LSTMs across all architectures.}
    \label{tab:shallowvsdeep}
    \begin{tabular}{ll|cccc}
    $D$ & & DCL & CausalBatch & Shallow D. & DCC \\ \hline
    \multirow{2}{*}{\cite{bockWEAROutdoorSports2024}}                   & 1-L & 73.64\% & 80.73\% & 81.98\% & 80.50\% \\
                                                                        & 2-L & 73.37\% & 75.15\% & 76.70\% & 78.46\% \\ \hline
    \multirow{2}{*}{\cite{schollWearablesWetLab2015}}                   & 1-L & 29.89\% & 33.63\% & 44.79\% & 46.52\% \\
                                                                        & 2-L & 26.42\% & 24.22\% & 36.00\% & 42.25\% \\ \hline
    \multirow{2}{*}{\cite{hoelzemannHangTimeHARBenchmark2023}}          & 1-L & 43.22\% & 47.79\% & 52.76\% & 53.27\% \\
                                                                        & 2-L & 40.80\% & 47.97\% & 49.58\% & 52.21\% \\ \hline
    \multirow{2}{*}{\cite{sztylerOnBodyLocalizationWearable2016}}       & 1-L & 73.64\% & 80.14\% & 81.27\% & 84.92\% \\
                                                                        & 2-L & 70.53\% & 68.66\% & 69.48\% & 89.50\% \\ \hline
    \multirow{2}{*}{\cite{roggenCollectingComplexActivity2010}}         & 1-L & 19.84\% & 29.92\% & 26.08\% & 45.06\% \\
                                                                        & 2-L & 29.41\% & 37.78\% & 38.05\% & 40.41\% \\ \hline
    \multirow{2}{*}{\cite{reyes-ortizTransitionAwareHumanActivity2016}} & 1-L & 56.55\% & 68.81\% & 72.66\% & 79.48\% \\
                                                                        & 2-L & 55.73\% & 66.56\% & 59.93\% & 71.29\% \\ \hline
    \multirow{2}{*}{\textit{Avg}}                                       & 1-L & 49.46\% & 56.84\% & 59.92\% & 64.96\% \\
                                                                        & 2-L & 49.37\% & 53.39\% & 54.96\% & 62.35\% \\
    \end{tabular}
    \vspace{-10pt}
\end{table}

\section{Limitations}
\label{sec:limitations}

Although not specifically tested within this work, architectural design choices of the DeepConvContext can be readily extended to other inertial-based models such as TinyHAR \cite{zhouTinyHARLightweightDeep2022} and Attend-and-Discriminate \cite{abedinAttendDiscriminateStateOfTheArt2021}. While our analysis evaluates only modified versions of the original DeepConvLSTM, we expect the observed performance improvements to generalize to other architectures, as they follow a common design pattern of combining convolutional feature extraction with some form of temporal modeling. Compared to the original DeepConvLSTM and the Shallow DeepConvLSTM, our architecture extends the original model design with additional modules and hence increases learnable parameters, yet, as shown in Table~\ref{tab:complexity}, has comparable throughput rate and model inference speed through the use of a dimensionality reduction layer. 
With HAR research and their applications using edge devices focus mostly on online activity recognition approaches, we did not expand on the offline activity recognition performances of DeepConvContext. Yet, as ablation experiments using bidirectional LSTMs within the DeepConvContext architecture show increased performance on HAR datasets, we expect DeepConvContext to be also suited for these purposes.

\section{Discussion \& Conclusions}
\label{sec:conclusion}

In this paper we introduced \textit{DeepConvContext}, a novel architecture for inertial-based activity recognition. Inspired by design choices in the vision-based Temporal Action Localization community, the architecture adopts a multi-scale approach that learns both intra- and inter-window temporal dependencies through separate model components to predict the label of a given sliding window. While our approach challenges the conventional definition of a batch by leaving temporal consistency among sliding windows in batches intact, our analysis revealed that these sequence-based batches are inherently more suitable for learning inter-window temporal features. 
Our benchmark analysis shows that DeepConvContext outperforms the standard DeepConvLSTM by up to 21\% and the Shallow DeepconvLSTM by on average 5\% in F1-score, while achieving the highest mAP across all six datasets -- improving of up to 18 points over the Shallow DeepConvLSTM, reflecting more coherent and less fragmented activity segment predictions. Unlike common post-hoc smoothing or filtering approaches (e.g., majority voting), which act only on already-made per-window decisions and can erode short but genuine transition activities, DeepConvContext learns coherent segments directly from the sensor data by shaping representations before a decision is made. Ablation experiments further reveal that LSTMs consistently outperform attention and transformer-based inter-window mechanisms across all benchmarks, suggesting that the sequential, locally-ordered inductive bias of LSTMs is better suited to HAR than the global, order-agnostic structure of self-attention. Context length ablations further reveal that DeepConvContext is robust to the choice of batch size, retaining competitive performance with as few as 25 windows (12.5 seconds of data).

With DeepConvContext, we present the HAR community with a novel method for processing windowed data and predicting activity labels. We directly address the well-known sliding window problem \cite{bullingTutorialHumanActivity2014} and align with the community's ongoing efforts to overcome it \cite{pellattCausalBatchSolvingComplexity2020, hiremathRoleContextLength2021}. By compressing inter-window communication to a fixed-size feature vector, our suggested changes do not come with sacrifices in efficiency and are easily extendable to other architectures. We thus hope that our proposed multi-scale approach will become an essential component within future works.

\begin{acks}
This project was supported by the ERC Consolidator Grant FORHUE (101044724), the DFG Project WASEDO (506589320) and the University of Siegen's OMNI cluster.
\end{acks}

\bibliographystyle{ACM-Reference-Format}
\balance
\bibliography{main}

\appendix

\end{document}